\documentclass{article}

\usepackage[english]{babel}

\usepackage[letterpaper,top=2cm,bottom=2cm,left=3cm,right=3cm,marginparwidth=1.75cm]{geometry}

\usepackage{multirow}
\usepackage{graphicx}
\usepackage{subcaption}
\usepackage{amsmath}
\usepackage{amsfonts}
\usepackage{amssymb}
\usepackage{amsthm}

\usepackage{algorithm}
\usepackage{algorithmic}

\usepackage{mathtools}  
\usepackage{bm}         
\usepackage{mathrsfs}
\newtheorem{definition}{Definition}[section]
\newtheorem{theorem}{Theorem}[section]
\newtheorem{lemma}{Lemma}[section]
\newtheorem{assumption}{Assumption}[section]
\newtheorem{remark}{Remark}[section]
\newtheorem{corollary}{Corollary}[section]
\usepackage{dsfont}

\usepackage{hyperref}   
\usepackage{cleveref}   

\usepackage{booktabs}
\usepackage{url}
\usepackage{xcolor}
\usepackage{cite}
\usepackage[T1]{fontenc}

\title{Training Task Reasoning LLM Agents for Multi-turn Task Planning via Single-turn Reinforcement Learning}
\author{Hanjiang Hu$^{1,3}$ ~ Changliu Liu$^{1}$ ~ Na Li$^{2,3}$ ~ Yebin Wang$^{3}$\\
$^{1}$Carnegie Mellon University ~ $^{2}$Harvard University \\ $^{3}$Mitsubishi Electric Research Laboratories (MERL) 
\thanks{
This work was done while Hanjiang Hu was a research intern and Na Li was a visiting scholar at Mitsubishi Electric Research Laboratories (MERL), Cambridge, MA 02139, USA. (email: {\tt\small hanjianghu@cmu.edu}). Yebin Wang is with the MERL. (email: \tt\small yebinwang@ieee.org).}
\thanks{Accepted by IEEE
Control Systems Letters (L-CSS)}
}

\date{} 
\begin{document}
\maketitle

\begin{abstract}
Large Language Models (LLMs) have demonstrated remarkable capabilities in knowledge acquisition, reasoning, and tool use, making them promising candidates for autonomous agent applications. However, training LLM agents for complex multi-turn task planning faces significant challenges, including sparse episode-wise rewards, credit assignment across long horizons, and the computational overhead of reinforcement learning in multi-turn interaction settings. To this end, this paper introduces a novel approach that transforms multi-turn task planning into single-turn task reasoning problems, enabling efficient policy optimization through Group Relative Policy Optimization (GRPO) with dense and verifiable reward from expert trajectories. \textcolor{black}{Our theoretical analysis shows that GRPO improvement on single-turn task reasoning results in a lower bound of the multi-turn success probability under the minimal turns, as well as the generalization to subtasks with shorter horizons.} Experimental evaluation on the complex task planning benchmark demonstrates that our 1.5B parameter model trained with single-turn GRPO achieves superior performance compared to larger baseline models up to 14B parameters, \textcolor{black}{with success rates of 70\% for long-horizon planning tasks.} 
\end{abstract}

\section{Introduction}

Large Language Models (LLMs) as autonomous agents are important in modern AI-based systems, which can perceive environments, reason about plans, and execute actions to interact with the environments \cite{luo2025large}.
Modern LLM agents demonstrate strong capabilities in knowledge integration, multi-step reasoning, and adaptive planning, as evidenced by their success in applications ranging from web search to robotic control \cite{xi2025rise,erdoganplan}. On top of these capabilities, prompt-based agentic frameworks \cite{yao2023react,shinn2023reflexion,zhou2024language} are proposed by integrating observation for environment state, reasoning based on augmented LLM with tools and memory, and action execution that interacts with the environment through structured interfaces as a series of \textit{single-turn} interactions with the environments. However, effort-costly prompt engineering is inevitable to build the LLM-based agent, and it is also computationally expensive for test-time scaling in the \textit{multi-turn} interaction with the environment \cite{guan2025evaluating,zhou2025sweet}. 

Therefore, training LLM agents through reinforcement learning (RL) for complex multi-turn task planning becomes a promising way to build effective agentic systems with low test-time cost \cite{wang2025ragen, feng2025group, dong2025agentic}. However, current RL approaches face critical challenges when applied to multi-turn interactions 
with the environment for LLMs \cite{cao2025skyrl,jin2025search,qi2024webrl,xi2025agentgym}. The reward of task completion  is too sparse for effective exploitation of LLM policy in the multi-turn settings. Besides, the credit assignment becomes particularly difficult in long-horizon planning problems, where determining which specific actions contribute to the eventual reward of success or failure is non-trivial \cite{feng2025group}. The computational complexity of multi-turn RL scales poorly with sequence length, making it impractical for training on complex tasks that may require dozens of sequential decision turns~\cite{peiyuan2024agile}.

In contrast, recent RL post-training successfully boosts LLM reasoning capability, particularly in math and coding domains based on on-policy methods, e.g., Group Relative Policy Optimization (GRPO) \cite{shao2024deepseekmath,guo2025deepseek}. However, these successes have primarily focused on single-turn reasoning tasks 
with verifiable reward, where the model generates a complete solution in one step and receives the reward directly based on the correctness of the final answer \cite{mroueh2025reinforcement}. The fundamental challenge lies in bridging the gap between single-turn reasoning capabilities and multi-turn task planning requirements, where LLM agents must be coherent along the interaction sequences while adapting to changing environmental states.

To this end, we aim to bridge the gap based on the insight that complex multi-turn task planning can be decomposed into a sequence of single-turn task reasoning problems, each requiring the agent to select the optimal next action given the current state.
This decomposition enables us to leverage the provably effective and efficient RL post-training with dense and verifiable reward based on expert trajectory, with theoretical guarantees of the multi-turn decision-making capabilities for long-horizon planning. We compare and validate the effectiveness and generalizability of the proposed method on the challenging long-horizon task planning benchmark. The contributions are summarized below.
\begin{itemize}
    \item \textcolor{black}{We provide a theoretical analysis bridging single-turn task reasoning and multi-turn task completion, proving that the success probability after GRPO can be lower-bounded in multi-turn planning scenarios. }
    \item We show empirical results on multiple tasks from the challenging {\scshape Robotouille} benchmark, where our 1.5B model trained with single-turn GRPO consistently outperforms 14B baselines with fewer completion steps.
    \item We both theoretically and empirically show the strong cross-task generalization capabilities of the GRPO-optimized LLM agents, where models trained on complex tasks successfully transfer to simpler subtasks.
\end{itemize}

\section{Problem Formulation}

We formulate the multi-turn task planning problem as two interconnected Markov Decision Processes (MDPs): a multi-turn MDP representing the complete long-horizon task planning and a single-turn MDP based on expert trajectories for training purposes.

\subsection{Multi-Turn Task Planning MDP}

Let $\mathcal{M} = (\mathcal{S}, \mathcal{A}, f, R, \mathcal{T}, s_0)$ be a finite-horizon MDP representing the multi-turn task planning problem, where $\mathcal{S}$ is the token state space representing the current task context and environment observation; $\mathcal{A}$ is the token action space of discrete possible single-step actions; $f: \mathcal{S} \times \mathcal{A} \rightarrow \mathcal{S}$ is the deterministic but unknown token state transition function; $R: \mathcal{S} \times \mathcal{A} \rightarrow \{0, 1\}$ is the binary sparse reward function indicating whether the task is completed; $\mathcal{T} \subset \mathbb{N}$ is the finite time horizon; $s_0 \in \mathcal{S}$ is the initial state.
Let $\pi: \mathcal{S} \rightarrow \mathcal{A}$ be a stochastic policy mapping token states to token action distributions. $\pi(a|s)$ denotes the probability of $a=\pi(s)$. A trajectory $\tau = [(s_0, a_0), (s_1, a_1), \ldots, (s_T, a_T)]$ is generated by following policy $\pi$ for $T$ steps, where $s_{i+1} = f(s_i, a_i)$. Then we define the successful trajectory induced by a stochastic policy as follows.

    
\begin{definition}[Successful trajectory by stochastic policy]
\label{def:episode_success}
Given the MDP $\mathcal{M} = (\mathcal{S}, \mathcal{A}, f, R, \mathcal{T}, s_0)$, a successful trajectory starting with $s_0^{s}=s_0$ is defined as:
\begin{align}
   \tau^{s}(T) = [(s_0^{s}, a_0^{s}), \ldots, (s_T^{s}, a_T^{s})], \ R(s_T^{s}, a_T^{s})=1,
\end{align}
where $s_{i+1}^{s} = f(s_i^{s},a_i^{s})$ for $i = 0,1,\dots, T-1$; $a_i^{s} \sim \pi(\cdot|s_{i}^{s})$ for $i = 0,1,2,\dots, T$ given the stochastic policy $\pi$.
\end{definition}

\subsection{Expert Trajectory with Unique Optimality}
On top of \Cref{def:episode_success}, we further define the expert trajectory generated by the expert policy with the unique optimality of minimal trajectory length.  \textcolor{black}{We adopt superscript ${GT}$ to denote ``Ground Truth'' for expert policy or trajectory.}
\begin{definition}[Expert trajectory by stochastic policy]
\label{def:expert_traj}
Given the MDP $\mathcal{M} = (\mathcal{S}, \mathcal{A}, f, R, \mathcal{T}, s_0)$, a trajectory $\tau^{GT}(T^{GT})$ from the stochastic policy  $\pi$ is defined as an expert trajectory if the following conditions hold:
\begin{enumerate}
    \item $\tau^{GT}(T^{GT})$ is a successful trajectory generated from stochastic policy the  $\pi$ in \Cref{def:episode_success}.
   \item For any successful trajectory $\tau(T)$ generated by stochastic policy $\pi$ with $R(s_T, a_T) = 1$, we have $T \geq T^{GT}$ and the equality holds if and only if $\tau=\tau^{GT}$.
\end{enumerate}
\end{definition}

Based on the expert trajectory $\tau^{GT}$ in \Cref{def:expert_traj}, we can further define the expert policy $\pi^{GT}$ as follows.
\begin{definition}[Expert policy from expert trajectory]
\label{def:expert_policy}
Given the MDP $\mathcal{M} = (\mathcal{S}, \mathcal{A}, f, R, \mathcal{T}, s_0)$ and an expert trajectory $\tau^{GT}(T^{GT})$, a policy $\pi^{GT}: \mathcal{S}\rightarrow \mathcal{A}$ is defined as the expert policy if it holds that
\begin{align}
    a_i^{GT} = \pi^{GT}(s_i^{GT}), \ \forall (s_i^{GT}, a_i^{GT}) \in \tau^{GT}(T^{GT}).
\end{align}
\end{definition}
\begin{remark}
     Even though expert trajectory $\tau^{GT}(T^{GT})$ is uniquely optimal with minimal length of turns $T^{GT}$, the induced expert policy $\pi^{GT}$ is not necessarily unique or deterministic, as there is no state-action
    constraint beyond expert trajectories $\tau^{GT}(T^{GT})$. \textcolor{black}{The unique optimal expert trajectory assumption ensures a well-defined single-step reward and serves as the foundation for the following results. We ensure this assumption in the experiment and further conduct an ablation comparison of robustness under non-expert noisy training trajectories.}
\end{remark}


\subsection{Single-Turn Task Reasoning MDP}
Even though multi-turn task planning MDP $\mathcal{M}$ is the conventional one in the control and RL community, it is challenging and non-trivial to directly train an LLM policy for $\mathcal{M}$ due to intractable computational overhead with exponentially increasing token length of trajectory rollouts in the multi-turn setting \cite{peiyuan2024agile,feng2025group}. Therefore, to enable efficient LLM policy learning 
, we construct an additional single-turn MDP $\mathcal{M}_S = (\mathcal{S}, \mathcal{A},\emptyset, r_{\pi^{GT}}, \mathit{1}, s_0)$ based on the expert trajectory $\tau^{GT}(T^{GT})$ in \Cref{def:expert_traj} and expert policy $\pi^{GT}$ in \Cref{def:expert_policy}. This single-turn MDP can be viewed as a bandit problem as follows.
$\mathcal{S}$ is the same token state space as in $\mathcal{M}$, $\mathcal{A}$ is the same token action space as in $\mathcal{M}$, $r_{\pi^{GT}}: \mathcal{S} \times \mathcal{A} \rightarrow \{0, 1\}$ is the binary reward function. There is no state transition function, which is denoted as $\emptyset$. The horizon $\mathit{1}$ is one-step in the single-turn setting. Initial state is the same as in $\mathcal{M}$.

\begin{definition}[Reward function from expert trajectory]
\label{def:stepwise_reward}
The binary reward function $r_{\pi^{GT}}$ for single-turn MDP $\mathcal{M}_S = (\mathcal{S}, \mathcal{A},\emptyset, r_{\pi^{GT}}, \mathit{1}, s^{GT})$ is defined as:
\begin{align}
    r_{\pi^{GT}}(s, a) = \mathds{1}\{a = \pi^{GT}(s)\} \in \{0, 1\},
\end{align}
where $\pi^{GT}$ is the expert policy defined in \Cref{def:expert_policy}.
\end{definition}

The single-turn MDP $\mathcal{M}_S$ is based on expert trajectories $\tau^{GT}$, which enables efficient policy optimization with dense reward in \Cref{def:stepwise_reward}. From the methodology section below, we will show that good performance on single-turn task reasoning (high reward $r$ in $\mathcal{M}_S$) will lead to successful long-horizon task planning (high reward $R$ in $\mathcal{M}$). 

\section{Methodology}

We first give the theoretical analysis of policy learning for single-turn MDP $\mathcal{M}_S$ based on expert trajectories.Then we define the success probability with the minimal turns for task planning MDP $\mathcal{M}$ and \textcolor{black}{show that policy improvement from $\mathcal{M}_S$ can give a lower bound of success probability in $\mathcal{M}$, for even unseen task planning. }

\subsection{GRPO for Single-Turn Task Reasoning}

GRPO is an on-policy model-free RL method that optimizes policies using verifiable rewards and replaces critic model with a group of multiple responses for a prompt to calculate a group-based advantage \cite{shao2024deepseekmath,guo2025deepseek}.
Based on the theoretical foundation of GRPO \cite{mroueh2025reinforcement},
we give the policy iteration and the success amplification
in the single-turn MDP $\mathcal{M}_S$ defined with \Cref{def:stepwise_reward}. 

\subsubsection{GRPO Policy Iteration Dynamics}

Let $\rho_Q$ be a distribution over state query prompts $s\in\mathcal{S}\sim \rho_Q$. For the single-turn MDP $\mathcal{M}_S$, we consider the GRPO optimization problem which adopts the relative advantage within a group of policy outputs as the critic while maintaining proximity to a reference policy through KL regularization.
The GRPO objective without clipping is formulated as:
\begin{align}
\max_{\pi} \mathbb{E}_{s \sim \rho_Q} \mathbb{E}_{a \sim \pi_{\text{old}}(\cdot|s)} \frac{\pi(a|s)}{\pi_{\text{old}}(a|s)} A(s, a) - \beta \text{KL}(\pi||\pi^{\text{ref}}),
\end{align}
where the advantage function $A$ based on the binary reward in \Cref{def:stepwise_reward} is given by:
\begin{align}
A_{\pi^{GT}}(s, a) = \frac{r_{\pi^{GT}}(s, a) - \mathbb{E}_{a' \sim \pi_{\text{old}}(\cdot|s)} r_{\pi^{GT}}(s, a')}{\sqrt{\text{Var}_{a' \sim \pi_{\text{old}}(\cdot|s)} r_{\pi^{GT}}(s, a')}}.
\end{align}
At iteration $n \geq 1$, the policy iteration of GRPO for $\mathcal{M}_S$ is,
\begin{align}
\label{eq:grpo_iter}
\pi_n = \arg\max_{\pi} \mathbb{E}_{s \sim \rho_Q} \bigg\{ \mathbb{E}_{a \sim \pi_{n-1}(\cdot|s)} \bigg[\frac{\pi_n(a|s)}{\pi_{n-1}(a|s)}
\frac{r_{\pi^{GT}}(s, a) - \mathbb{E}_{a' \sim \pi_{n-1}(\cdot|s)} r_{\pi^{GT}}(s, a')}{\sqrt{\text{Var}_{a' \sim \pi_{n-1}(\cdot|s)} r_{\pi^{GT}}(s, a')}}\bigg] - \beta \text{KL}(\pi_n||\pi^{\text{ref}}) \bigg\}, 
\end{align}
where initial policy is the reference policy $\pi_0 = \pi^\text{ref}$.


\subsubsection{GRPO Single-Turn Optimality and Success Amplification}

Following the \Cref{thm:grpo_amplify} from \cite{mroueh2025reinforcement} below, we can have the optimality guarantee for GRPO iteration in \Cref{eq:grpo_iter} in our single-turn task reasoning setting.
\begin{theorem}[Theorem 3 from \cite{mroueh2025reinforcement}]
\label{thm:grpo_amplify}
Let $\pi^*$ be the fixed point of GRPO-optimized policy obtained from the policy iteration dynamics of \Cref{eq:grpo_iter} and $\pi^{\text{ref}}$ be the initial reference policy. Denote the success probability of the reference policy as $p^{\text{ref}} = \mathbb{E}_{a \sim \pi^{\text{ref}}(\cdot|s^{GT})} [\mathds{1}\{a = \pi^{GT}(s^{GT})\}]$ and optimized policy as $p^{*} = \mathbb{E}_{a \sim \pi^{*}(\cdot|s^{GT})} [\mathds{1}\{a = \pi^{GT}(s^{GT})\}]$.   It holds that $p^* > p^{\text{ref}}$.
\end{theorem}
\textcolor{black}{
\begin{proof}
The optimal policy at iteration $n$ of the GRPO dynamics in \Cref{eq:grpo_iter} satisfies 
$
\pi_n(a|s) \propto \pi_{\text{ref}}(a|s) \exp\left(\frac{1}{\beta} A_{\pi^{GT}}(s,a)\right),$
where the advantage function $A_{\pi^{GT}}$ upweights successful actions ($r_{\pi^{GT}}(s,a)=1$) and downweights failures ($r_{\pi^{GT}}(s,a)=0$). The variance normalization in $A_{\pi^{GT}}$ ensures positive advantages for successes and negative advantages for failures. At the fixed point $\pi^*$, this means $\exp(\frac{1}{\beta}A_{\pi^{GT}}) > 1$ for correct actions and $\exp(\frac{1}{\beta}A_{\pi^{GT}}) < 1$ for incorrect ones, systematically shifting probability mass toward successful outputs and yielding $p^* > p_{\text{ref}}$.  Check Appendix C.1 of \cite{mroueh2025reinforcement} for full proof.
\end{proof}
}
\begin{corollary}[GRPO single-Turn optimality]
\label{thm:grpo_single_turn}
    For the fixed point of GRPO-optimized policy $\pi^*$  and  the  reference policy $\pi^{\text{ref}}$, the following holds for any state $s$ from the expert trajectory $\tau^{GT}$,
\begin{align}
\label{eq:singleturn_amplify}
    \mathbb{E}_{a^* \sim \pi^*(\cdot|s)} [r_{\pi^{GT}}(s, a^*)] \geq \mathbb{E}_{a^{\text{ref}} \sim \pi^{\text{ref}}(\cdot|s)} [r_{\pi^{GT}}(s, a^{\text{ref}})].
\end{align}
\end{corollary}
\begin{proof}
It holds based on the single-turn reward formulation in \Cref{def:stepwise_reward} based on \Cref{thm:grpo_amplify}.
\end{proof}

\Cref{thm:grpo_single_turn} establishes that GRPO training on the single-turn MDP $\mathcal{M}_S$ with expert trajectories leads to policies that achieve higher success rates than the reference policy before GRPO training, providing the foundation for analyzing multi-turn performance improvements.

\subsection{Success Probability for Multi-Turn Task Planning}

To analyze how GRPO improvements on the single-turn MDP $\mathcal{M}_S$ can  result in performance gains in the multi-turn MDP $\mathcal{M}$, we introduce the success probability with the minimal turns that captures the probability of achieving the optimal expert trajectory, by assuming the existence of minimal completion steps given any state below.

\begin{definition}[Minimal turns for task completion]
\label{def:minimal_steps}
For any state $s \in \mathcal{S}$ in the multi-turn MDP $\mathcal{M}$, denote the unique minimal length of turns as $T^*(s) \geq 0$ among all successful trajectories defined in \Cref{def:episode_success}  from $s$, i.e., $T^*(s) := \min\{T \geq 0 : \exists \tau(T) \text{ s.t. } R(s_{T}, a_{T}) = 1, s_0=s\}.$
\end{definition}

$T^*(s)$ exists uniquely based on the existence and uniqueness of the expert trajectory in \Cref{def:expert_traj} by taking any states as initial states. We then define the probability of the successful trajectory with minimal steps.
\begin{definition}[Success probability with minimal steps]
\label{def:value_function}
With $T^*(s)$ in \Cref{def:minimal_steps}, the probability of achieving the task planning goal in the minimal number of steps starting from state $s_t$ at time $t$ under policy $\pi$ in the multi-turn MDP $\mathcal{M}$ is defined as,
\begin{align}
\label{eq:multi-turn-success}
    P_t^\pi(s_t) = \mathbb{P}_{\pi}\left(R(s_{t+T^*(s_t)}, a_{t+T^*(s_t)}) = 1 | s_t\right).
\end{align}
\end{definition}

We then present the recursion condition for the success probability in \Cref{def:value_function} based on the reward function from the expert trajectory in \Cref{def:stepwise_reward}.

\begin{theorem}[Recursion equation of success probability]
\label{thm:bellman_recursion}
The success probability defined in \Cref{def:value_function} satisfies the following recursion condition,
\begin{align}
    P_t^\pi(s_t) = \mathbb{E}_{a \sim \pi(\cdot|s_t)} [r_{\pi^{GT}}(s_t, a) \cdot P_{t+1}^\pi(s_{t+1})],
\end{align}
where $\pi^{GT}$ is the expert policy in \Cref{def:expert_policy}  by expert trajectory with $P_{t+T^*(s_t)}^\pi(s_{t+T^*(s_t)}) = 1$ for any state $s_t$.
\end{theorem}

\begin{proof}
For any state $s_t$ as the initial state, based on \Cref{def:expert_policy}, $\pi^{GT}$ is induced by the expert trajectory $\tau^{GT} =[(s_t, a_t), \dots, (s_{T^{GT}}, a_{T^{GT}})], s_{k+1} = f(s_k, \pi^{GT}(s_k)), k=t,t+1,\dots, T^{GT}-1$ with minimal turns of task completion. By $P_{t+T^*(s_t)}^\pi(s_{t+T^*(s_t)}) = 1$, we have $R(s_{t+T^*(s_t)}, a_{t+T^*(s_t)}) = 1$. Combining the uniqueness of expert trajectory in \Cref{def:expert_traj} and  minimal completion steps $T^*(s)$ in  \Cref{def:minimal_steps}, it holds that 
\begin{align}
\label{eq:same_turn_condition}
    T^*(s_{T^{GT}})=0, T^{GT} = T^*(s_k) + k, k=t,\dots, T^{GT}.
\end{align}
For any state that deviates from the expert trajectory $s_{k+1}' \neq f(s_k, \pi^{GT}(s_k))$, by the minimal completion turns, 
\begin{align}
\label{eq:min_turn_condition}
    T^*(s_{k+1}') > T^*(s_{k}) - 1, \forall k = t, t+1, \dots, T^{GT} - 1.
\end{align}
By the law of total expectation over the next action $t+1$:
\begin{align}
\label{eq:total_exp}
&P_t^\pi(s_t) = \mathbb{P}_{\pi}(R(s_{t+T^*(s_t)}, a_{t+T^*(s_t)}) = 1 | s_t) \\
&= \mathbb{E}_{a \sim \pi(\cdot|s_t)}[\mathbb{P}_{\pi}(R(s_{t+T^*(s_t)}, a_{t+T^*(s_t)}) = 1 | s_t, a)]. \notag
\end{align}
For $a \sim \pi(\cdot|s_t)$, if  $a \neq \pi^{GT}(s_t),i.e., r_{\pi^{GT}}(s_t, a) = 0$, we have $s_{t+1}' \neq f(s_t, \pi(s_t))$. Then based on \Cref{eq:min_turn_condition}, it takes strictly more than $T^*(s_{t}) - 1$ steps for completion, i.e., 
\begin{align}
\label{eq:nonexpert_nextstep}
    \mathbb{P}_{\pi}(R(s_{t+1+T^*(s_{t})-1}, a_{t+1+T^*(s_{t})-1}) = 1 | s_{t+1}')=0.
\end{align}
 If $a = \pi^{GT}(s_t),  i.e., r_{\pi^{GT}}(s_t, a) = 1, s_{t+1}=f(s_t, a)$, then based on \Cref{eq:same_turn_condition}, we have 
 \begin{align}
 \label{eq:expert_nextstep}
    &\mathbb{P}_{\pi}(R(s_{t+T^*(s_t)}, a_{t+T^*(s_t)}) = 1 | s_t, a) \\=&\mathbb{P}_{\pi}(R(s_{t+1+T^*(s_{t+1})}, a_{t+1+T^*(s_{t+1})}) = 1 | s_{t+1}) \notag\\=& P_{t+1}^\pi(s_{t+1}).\notag
\end{align}
With the values of $r_{\pi^{GT}}(s_t, a)\in\{0,1\}$ and combining \Cref{eq:total_exp,eq:nonexpert_nextstep,eq:expert_nextstep}, it holds that
 \begin{align}
    P_t^\pi(s_t) = \mathbb{E}_{a \sim \pi(\cdot|s_t)} [r_{\pi^{GT}}(s_t, a) \cdot P_{t+1}^\pi(s_{t+1})],
\end{align}
which concludes the proof.
\end{proof}

Please note that the recursion equation in \Cref{thm:bellman_recursion} is not the same as Bellman recursion in value function iteration, and the key difference lies in that the multiplication of the binary single-turn reward $r_{\pi^{GT}}(s_t, a)$ is involved in the recursion of expectation, instead of addition. This recursion bridges the final multi-turn success in \Cref{eq:multi-turn-success} with single-turn reward improvements at each step in \Cref{eq:singleturn_amplify}, \textcolor{black}{which is crucial to the following analysis on how single-turn GRPO gives a lower bound of success probability of the multi-turn task planning.}

\subsection{\textcolor{black}{Lower Bound of Multi-Turn Success Probability}}

\textcolor{black}{We now analyze how GRPO improvements on single-turn task reasoning in $\mathcal{M}_S$ will give a lower bound of success probability in the multi-turn task planning MDP $\mathcal{M}$. First, we assume the bounded Total Variation (TV) distance of actions between the proximal states in the training and test data. Then we show the lower bound of single-turn GRPO performance as Lemma \ref{lem:grpo_generalization}. 
\begin{assumption}[Bounded Policy with State Proximity]
\label{ass:grpo_generalization}
For any execution state $s \in \mathcal{S}_{\text{exec}}$ encountered during policy deployment, there exists a training state $s' \in \mathcal{S}_{\text{train}}$ with the same expert action $\pi^{GT}(s') = \pi^{GT}(s)$ such that for any policy $\pi$, $\|\pi(\cdot|s) - \pi(\cdot|s')\|_{\text{TV}} \leq \delta$,
where $\delta > 0$ is a bounded constant and $\|\cdot\|_{\text{TV}}$ denotes the TV distance.
\end{assumption}
\begin{lemma}[Lower bound of GRPO policy deployment]
\label{lem:grpo_generalization}
Under Assumption~\ref{ass:grpo_generalization}, for any execution state $s \in \mathcal{S}_{\text{exec}}$,
    $\mathbb{E}_{a \sim \pi^*(\cdot|s)}[r_{\pi^{GT}}(s,a)] \geq \mathbb{E}_{a \sim \pi^{\text{ref}}(\cdot|s)}[r_{\pi^{GT}}(s,a)] - 4\delta,$
where $\pi^*$ is the GRPO-optimized policy and $\pi^{\text{ref}}$ is the reference policy.
\end{lemma}
\begin{proof}
Let $s \in \mathcal{S}_{\text{exec}}$. By Assumption~\ref{ass:grpo_generalization}, there exists $s' \in \mathcal{S}_{\text{train}}$ such that $\|\pi^*(\cdot|s) - \pi^*(\cdot|s')\|_{\text{TV}} \leq \delta$, s$\|\pi^{\text{ref}}(\cdot|s) - \pi^{\text{ref}}(\cdot|s')\|_{\text{TV}} \leq \delta$, and $\pi^{GT}(s') = \pi^{GT}(s)$. With $r_{\pi^{GT}}(s,a)=r_{\pi^{GT}}(s',a) \in \{0,1\}$, for any policy $\pi$ and states $s, s'$ with $\|\pi(\cdot|s) - \pi(\cdot|s')\|_{\text{TV}} \leq \delta$, we have:
\begin{align}
\label{eq:tv_bound}
    &\left|\mathbb{E}_{a \sim \pi(\cdot|s)}[r_{\pi^{GT}}(s,a)] - \mathbb{E}_{a \sim \pi(\cdot|s')}[r_{\pi^{GT}}(s',a)]\right|\\
    &\leq \sum_{a \in \mathcal{A}} |\pi(a|s) - \pi(a|s')| \cdot|r_{\pi^{GT}}(s,a)|\notag\\ &\leq \sum_{a \in \mathcal{A}} |\pi(a|s) - \pi(a|s')| 
    = 2\|\pi(\cdot|s) - \pi(\cdot|s')\|_{\text{TV}} \leq 2\delta.\notag
\end{align}
By Corollary \ref{thm:grpo_single_turn}, since $s' \in \mathcal{S}_{\text{train}}$:
\begin{align}
    \mathbb{E}_{a \sim \pi^*(\cdot|s')}[r_{\pi^{GT}}(s',a)] 
    &\geq \mathbb{E}_{a \sim \pi^{\text{ref}}(\cdot|s')}[r_{\pi^{GT}}(s',a)]. \label{eq:grpo_train}
\end{align}
Applying \eqref{eq:tv_bound} to $\pi^{\text{ref}},\pi^*$ and combining  \eqref{eq:grpo_train} will complete the proof.
\end{proof}
Finally, the step-dependent lower bound of success probability is shown below, which becomes tighter as the time step is closer to the minimal length of turns $T^{GT}$ of expert trajectories. It is consistent with the intuition that the task is more likely to complete from the middle step compared to initial step.
\begin{theorem}[Lower bound of GRPO success probability]
\label{thm:value_improvement}
Let $\pi^*$ be the fixed point of the GRPO-optimized policy from the policy iteration dynamics of \Cref{eq:grpo_iter} and $\pi^{\text{ref}}$ be the initial reference policy. Under Assumption~\ref{ass:grpo_generalization} and \Cref{eq:singleturn_amplify}, it  holds for any given state $s_t$ at step $t$ that,
\begin{align}
    P_t^{\pi^*}(s_t) \geq P_t^{\pi^{\text{ref}}}(s_t) - (T^{GT}-t) \cdot 4\delta,
\end{align}
\end{theorem}
\begin{proof}
By the existence and uniqueness of the expert trajectory in \Cref{def:expert_traj}, denote the expert trajectory starting from $s_t$ as $\tau^{GT} =[(s_t, a_t), \dots, (s_{T^{GT}}, a_{T^{GT}})]$ with minimal turns of task completion $T^*(s_t) = T^{GT}-t$ in \Cref{def:minimal_steps}.
We prove by backward induction on the length of remaining turns to task completion. In the final case when $t=T^{GT}$, we have $P_t^{\pi^*}(s_t) = P_t^{\pi^{\text{ref}}}(s_t) = 1$ since the task is already completed.
 Now for the inductive case at step $t+1, t+1\leq T^{GT}$, we assume $P_{t+1}^{\pi^*}(s') \geq P_{t+1}^{\pi^{\text{ref}}}(s') - (T^{GT}-t-1) \cdot 4\delta$ for all $s' \in \mathcal{S}$. Then for any state $s$ at step $t$, by \Cref{thm:bellman_recursion}:
\begin{align}
&P_t^{\pi^*}(s)
= \mathbb{E}_{a \sim \pi^*(\cdot|s)} [r_{\pi^{GT}}(s,a) \cdot P_{t+1}^{\pi^*}(f(s,a))] \\
&\geq \mathbb{E}_{a \sim \pi^*} [r_{\pi^{GT}}(s,a) \{P_{t+1}^{\pi^{\text{ref}}}(f(s,a)) - (T^{GT}-t-1) 4\delta\}] \label{eq:ind_hyp}\\
&= \mathbb{E}_{a \sim \pi^*(\cdot|s)} [r_{\pi^{GT}}(s,a) \cdot P_{t+1}^{\pi^{\text{ref}}}(f(s,a))]  - (T^{GT}-t-1) \cdot 4\delta \cdot \mathbb{E}_{a \sim \pi^*(\cdot|s)} [r_{\pi^{GT}}(s,a)] \\
&\geq \mathbb{E}_{a \sim \pi^*(\cdot|s)} [r_{\pi^{GT}}(s,a) \cdot P_{t+1}^{\pi^{\text{ref}}}(f(s,a))]  - (T^{GT}-t-1) \cdot 4\delta \label{eq:bound_r}\\
&\geq \mathbb{E}_{a \sim \pi^{\text{ref}}(\cdot|s)} [r_{\pi^{GT}}(s,a) \cdot P_{t+1}^{\pi^{\text{ref}}}(f(s,a))]  - 4\delta - (T^{GT}-t-1) \cdot 4\delta \label{eq:apply_lem}\\
&= P_t^{\pi^{\text{ref}}}(s) - (T^{GT}-t) \cdot 4\delta, \label{eq:recursion_ref}
\end{align}
where \eqref{eq:ind_hyp} applies the inductive hypothesis, \eqref{eq:bound_r} uses $\mathbb{E}_{a \sim \pi^*(\cdot|s)} [r_{\pi^{GT}}(s,a)] \leq 1$, \eqref{eq:apply_lem} applies Lemma~\ref{lem:grpo_generalization} (noting that since $0 \leq P_{t+1}^{\pi^{\text{ref}}}(s') \leq 1$, the weighted expectation difference is bounded by the marginal difference), and \eqref{eq:recursion_ref} uses \Cref{thm:bellman_recursion} for $\pi^{\text{ref}}$.
\end{proof}
}

Furthermore, we show that the policy trained with expert trajectories with single-turn GRPO can also generalize to subtask defined below.
\begin{definition}[Subtask  and its success probability]
\label{def:subtask_completion}
Given the expert trajectory $\tau^{GT}(T^{GT})=[(s_0, a_0), \dots, (s_{T^{GT}}, a_{T^{GT}})]$ in \Cref{def:expert_traj} for the full task, a subtask is defined by a completion step $k^* < T^{GT}$ where the subtask is completed upon reaching $(s_{k^*}, a_{k^*})$, i.e., $R_{\text{sub}}(s_{k^*}, a_{k^*})=1$. The success probability with the fewest turns for subtask at $s_t, t\leq k^*$ is defined as,
\begin{align}
    P_t^{\text{sub}, \pi}(s_t) = \mathbb{P}_{\pi}\left(R_{\text{sub}}(s_{k^*}, a_{k^*}) =1 | s_t\right),
\end{align}
which is the probability of achieving subtask  goal in the minimal number of steps starting from state $s_t$ at time $t$ under policy $\pi$.
\end{definition}

Based on the success probability of subtask, we present the following corollary of \Cref{thm:value_improvement}.
\begin{corollary}[GRPO generalization to subtasks]
\label{cor:grpo_subtask}
For any subtask defined by completion step $k^* < T^{GT}$ in \Cref{def:subtask_completion}, the GRPO-optimized policy $\pi^*$ achieves better performance on the subtask than the reference policy $\pi^{\text{ref}}$ for any given state $s_t$ at step $t\leq k^*$, \textcolor{black}{i.e.,
$P_t^{\text{sub}, \pi^*}(s_t) \geq P_t^{\text{sub}, \pi^{\text{ref}}}(s_t)- (k^*-t) \cdot 4\delta$.}
\end{corollary}

\begin{proof}
Since the subtask uses the same expert trajectory prefix as the full task, Assumption \ref{ass:grpo_generalization} also holds for subtask goal with the same single-turn  reward function $r_{\pi^{GT}}(s,a)$. Besides,  the recursion condition of $P_t^{\text{sub}, \pi}(s_t)$ also applies. Therefore, following the proof of \Cref{thm:value_improvement}, by letting $T^{GT}=k^*$, $\pi^*$ achieves better success probability than $\pi^{\text{ref}}$ over $P_t^{\text{sub}}(s_t)$ for any given state $s_t$ at step $t\leq k^*$.
\end{proof}

\section{Experiments}
In the experiments, we aim to answer the following two questions empirically: \textit{1) How is the performance of multi-turn task planning after GRPO over single-turn task reasoning? \textcolor{black}{2) Are GRPO-trained agents generalizable to unseen planning tasks and robust to noisy non-expert training trajectories?}} We answer the first question in \Cref{sec:exp_compare} and the second one in \Cref{sec:exp_generalization}.
Prior to the answers, we first introduce the experimental setup of environment, model training and evaluation metrics.

\subsection{Experimental Setup}

\paragraph{Environment and Tasks}
We evaluate our approach on the {\scshape Robotouille} benchmark \cite{gonzalezrobotouille}, a challenging multi-turn task planning environment designed for testing LLM agents' reasoning capabilities in cooking scenarios. {\scshape Robotouille} provides a deterministic long-horizon complex environment where agents must coordinate multiple actions across different kitchen stations to complete recipes. \textcolor{black}{We focus on three progressively challenging cooking tasks with different maximum steps: (1) \textit{Burger} - at most 10 steps; (2) \textit{Cheese Burger} - at most 15 steps; and (3) \textit{Double Cheese Burger} - at most 23 steps. These tasks were selected to provide increasing complexity in both planning horizon and reasoning requirements while sharing the in-distribution states and actions, allowing us to evaluate our method's scalability and generalization across different task difficulties.}

\begin{table}[htbp]
\centering
\caption{Performance Comparison on Different Tasks. Best results are in \textbf{bold}. ``---'' indicates no successful trajectories.}
\label{tab:cheese_sandwich}
\resizebox{0.8\textwidth}{!}{
\textcolor{black}{
\begin{tabular}{cccc}
\toprule
Results of Burger & SR ($\uparrow$) & ASAT ($\downarrow$) & ASST ($\downarrow$)\\
\midrule
\textcolor{gray}{Llama3-70b (expert)} & \textcolor{gray}{0.53$\pm$0.15} & \textcolor{gray}{12.13$\pm$0.57} & \textcolor{gray}{9.53$\pm$0.50} \\
Qwen2.5-7b & 0.07$\pm$0.06 & 14.97$\pm$0.06 & 14.0$\pm$0.0 \\
Qwen2.5-14b & 0.10$\pm$0.10 & 14.87$\pm$0.15 & 13.67$\pm$0.29 \\
Qwen1.5b (w/ SFT) & 0.30$\pm$0.10 & 12.37$\pm$1.17 & 8.64$\pm$0.13 \\
{Qwen1.5b (w/ SFT \& GRPO)} & \textbf{0.73$\pm$0.06} & \textbf{10.07$\pm$0.25} & \textbf{8.18$\pm$0.16} \\
\midrule
Results of Cheeseburger & SR ($\uparrow$) & ASAT ($\downarrow$) & ASST ($\downarrow$)\\
\midrule
\textcolor{gray}{Llama3-70b (expert)} & \textcolor{gray}{0.73$\pm$0.06} & \textcolor{gray}{17.23$\pm$0.38} & \textcolor{gray}{15.12$\pm$0.13} \\
Qwen2.5-7b & 0.0$\pm$0.0 & 23.0$\pm$0.0 & --- \\
Qwen2.5-14b & 0.10$\pm$0.10 & 22.63$\pm$0.32 & 19.33$\pm$1.15 \\
Qwen1.5b (w/ SFT) & 0.57$\pm$0.06 & 16.97$\pm$1.36 & 14.73$\pm$0.49 \\
{Qwen1.5b (w/ SFT \& GRPO)} & \textbf{0.70$\pm$0.00} & \textbf{15.53$\pm$0.23} & \textbf{12.30$\pm$0.35} \\
\midrule
Results of Double Cheeseburger & SR ($\uparrow$) & ASAT ($\downarrow$) & ASST ($\downarrow$)\\
\midrule
\textcolor{gray}{Llama3-70b (expert)} & \textcolor{gray}{0.20$\pm$0.10} & \textcolor{gray}{32.50$\pm$0.89} & \textcolor{gray}{21.77$\pm$2.25} \\
Qwen2.5-7b & 0.0$\pm$0.0 & 35.0$\pm$0.0 & --- \\
Qwen2.5-14b & 0.0$\pm$0.0 & 35.0$\pm$0.0 & --- \\
Qwen1.5b (w/ SFT) & 0.13$\pm$0.06 & 22.87$\pm$0.85 & 27.17$\pm$2.75 \\
{Qwen1.5b (w/ SFT \& GRPO)} & \textbf{0.3$\pm$0.0} & \textbf{30.5$\pm$0.0} & \textbf{20.0$\pm$0.0} \\
\bottomrule
\end{tabular}
}
}
\end{table}

\begin{table}[htbp]
\centering
\caption{\textcolor{black}{Performance comparison on Burger task with different RL post-training methods. Best results are in \textbf{bold}.}}
\label{tab:rl_training}
\resizebox{0.8\textwidth}{!}{
\textcolor{black}{
\begin{tabular}{cccc}
\toprule
RL post-training methods & SR ($\uparrow$) & ASAT ($\downarrow$) & ASST ($\downarrow$)\\
\midrule
SFT \& PPO  & 0.60$\pm$0.10 & 11.40$\pm$0.61 & 8.53$\pm$0.29 \\
SFT \& RLOO & 0.70$\pm$0.00 & 10.37$\pm$0.06 & 8.38$\pm$0.08 \\
SFT \& REINFORCE++ & 0.67$\pm$0.12 & 10.80$\pm$1.11 & 8.30$\pm$0.05 \\
SFT \& GRPO & \textbf{0.73$\pm$0.06} & \textbf{10.07$\pm$0.25} & \textbf{8.18$\pm$0.16}\\ 
\bottomrule
\end{tabular}
}
}
\end{table}

\begin{table}[htbp]
\centering
\caption{\textcolor{black}{Cross-task generalization of SR and ASAT.}}
\label{tab:success_generalization}
\resizebox{0.8\textwidth}{!}{
\textcolor{black}{
\begin{tabular}{cccc}
\toprule
\begin{tabular}[c]{@{}c@{}} Evaluation Tasks ($\rightarrow$)\\ Model Training Tasks ($\downarrow$)\end{tabular}     &  \begin{tabular}[c]{@{}c@{}}Burger SR / \\ ASAT\end{tabular}  & \begin{tabular}[c]{@{}c@{}}Cheese Burger \\ SR / ASAT\end{tabular} & \begin{tabular}[c]{@{}c@{}}Double Cheese\\ Burger SR / ASAT\end{tabular} \\
\midrule
Burger  &  \begin{tabular}[c]{@{}c@{}}0.73$\pm$0.06 / \\10.07$\pm$0.25\end{tabular}  & \begin{tabular}[c]{@{}c@{}}0.0$\pm$0.0 / \\ 23.0$\pm$0.0\end{tabular} & \begin{tabular}[c]{@{}c@{}}0.0$\pm$0.0 / \\ 35.0$\pm$0.0\end{tabular} \\ \cmidrule{2-4}
Cheese Burger  & \begin{tabular}[c]{@{}c@{}}0.60$\pm$0.10 / \\10.90$\pm$0.17\end{tabular} & \begin{tabular}[c]{@{}c@{}}0.7$\pm$0.0 / \\ 15.50$\pm$0.23\end{tabular} & \begin{tabular}[c]{@{}c@{}}0.0$\pm$0.0 / \\ 35.0$\pm$0.0\end{tabular} \\ \cmidrule{2-4}
Double Cheese Burger  & \begin{tabular}[c]{@{}c@{}}0.5$\pm$0.0 / \\11.60$\pm$0.10\end{tabular} & \begin{tabular}[c]{@{}c@{}}0.37$\pm$0.06 / \\19.13$\pm$0.55\end{tabular}  & \begin{tabular}[c]{@{}c@{}}0.3$\pm$0.0 / \\30.5$\pm$0.0\end{tabular}  \\
\bottomrule
\end{tabular}
}
}
\end{table}

\begin{table}[htbp]
\centering
\caption{\textcolor{black}{Performance comparison on Burger task with different ratio of noisy trajectories. Best results are in \textbf{bold}.}}
\label{tab:noise_ratio}
\resizebox{0.8\textwidth}{!}{
\textcolor{black}{
\begin{tabular}{cccc}
\toprule
Ratio of noisy trajectories & SR ($\uparrow$) & ASAT ($\downarrow$) & ASST ($\downarrow$)\\
\midrule
Noisy, 50\%  & 0.50$\pm$0.10 & 11.90$\pm$0.26 & 8.23$\pm$0.21 \\
Noisy, 30\%  & 0.56$\pm$0.06 & 11.43$\pm$0.55 & 8.23$\pm$0.21 \\
Noisy, 10\%  & 0.67$\pm$0.06 & 10.47$\pm$0.29 & 8.19$\pm$0.17 \\
Clean, 0\% & \textbf{0.73$\pm$0.06} & \textbf{10.07$\pm$0.25} & \textbf{8.18$\pm$0.16}\\
\bottomrule
\end{tabular}
}
}
\end{table}

\paragraph{Data Collection and Model Training}
\textcolor{black}{Expert trajectories are collected using rejection sampling and minimum-step trajectory filtering from Llama3.3-70B-Instruct \cite{dubey2024llama},} which serves as our expert policy $\pi^{GT}$. To ensure diversity and robustness, we generate expert trajectories across varied kitchen layouts and ingredient placements, creating approximately 100 successful trajectories per task type. \textcolor{black}{For the ablation of training on noisy non-expert trajectories, we randomly choose 10\%, 30\%, 50\% of the states along all trajectories that are perturbed to have sub-optimal actions but keep the task completion successful.}
The state-action tokens follow the ReAct agentic framework \cite{yao2023react}, where each action $a_i^{GT}$ consists of structured reasoning followed by a specific action selection from the valid action list indicated from state tokens. 
ReAct format and single-turn correctness reward of \Cref{def:stepwise_reward} are weighted by 0.1 and 0.9, respectively.
Our training pipeline consists of two sequential post-training stages using the VeRL  implementation \cite{sheng2025hybridflow} of RL post-training. First, we perform supervised fine-tuning (SFT) on Qwen2.5-1.5B-Instruct \cite{qwen2.5} using the collected expert trajectories to establish a strong initialization. 
The SFT takes 8 epochs with batch size of 8 and the RL post-training uses a batch size of 256, learning rate of $1 \times 10^{-6}$, and KL regularization coefficient $\beta = 0.001$ for 50 epochs.



\paragraph{Evaluation Metrics and Baselines}
We evaluate all models on 10 unseen kitchen layouts per task type, ensuring that the test environments differ from training data in terms of object positions, kitchen topology, and ingredient availability. The timeout lengths of attempt turns are 15, 23, 35 for tasks of \textit{Burger, Cheese Burger, Double Cheese Burger}, respectively.
We assess the model  performance of effectiveness and efficiency using three key metrics: (1) \textit{Success Rate (SR)} - the proportion of episodes where the agent successfully completes the task within the step timeout limit; (2) \textit{Average Steps of All Trajectories (ASAT)} 
- the mean number of steps taken across all episodes, including timeout failures; and (3) \textit{Average Steps of Successful Trajectories (ASST)}
- the mean number of steps for successfully completed episodes only. \textcolor{black}{The temperature is set to be 0.7 and all values are in \textit{Mean $\pm$ Standard Deviation} from 3 random trials.
We compare our approach against prompt-based larger agents (7B, 14B models of Qwen2.5), 1.5B models only with SFT, and 1.5B models with SFT and other RL post-training methods (PPO \cite{schulman2017proximal}, RLOO \cite{ahmadian2024back}, REINFORCE++ \cite{hu2025reinforce++}).}

\textcolor{black}{
\subsection{Results of Task Planning with Single-turn GRPO}
\label{sec:exp_compare}
In \Cref{tab:cheese_sandwich}, the models with SFT and GRPO consistently outperform much larger-size baseline models and perform better or keep on par with expert policy. 
The SFT-alone model provides limited success compared to the one followed by GRPO, validating the effectiveness of single-turn RL for task reasoning in \Cref{thm:value_improvement}. 
Beyond success rates, the model with GRPO post-training consistently achieves lower average step counts for successful episodes, 
demonstrating that GRPO learns more optimal policies with minimal steps of task completion in \Cref{def:value_function}. 
}

\textcolor{black}{
From \Cref{tab:rl_training}, the proposed GRPO-based approach achieves the best performance in all metrics compared to other RL post-training methods. The PPO-based agent performs the worst due to its reliance on a separate critic network for advantage estimation, which introduces additional approximation errors. Compared to critic-free RL methods like RLOO and REINFORCE++,  group-relative advantage computation of GRPO provides more stable gradient estimates by normalizing within sampled responses.
}

\textcolor{black}{
\subsection{Ablation Study and Generalization Analysis}
\label{sec:exp_generalization}
\paragraph{Cross-task Generalization Analysis}
To evaluate generalizability, we train agents on individual tasks and assess zero-shot performance across all tasks. The results in \Cref{tab:success_generalization} empirically validate \Cref{cor:grpo_subtask}: agents trained on complex tasks generalize effectively to simpler ones, while the reverse fails. The Double Cheese Burger-trained agent achieves non-zero success rates on all tasks, whereas Burger-trained and Cheese Burger-trained agents completely fail on more complex task of Double Cheese Burger. 
 These findings demonstrate that our single-turn GRPO approach enables meaningful cross-task generalization from complex to simpler tasks, aligning with our theoretical predictions, though highlighting that complex reasoning capabilities cannot be readily acquired from simpler task training alone.
\paragraph{Robustness to Noisy Non-expert Trajectories}
To assess robustness to imperfect demonstrations, we evaluate the Burger task with varying ratios of noisy trajectories. Table \ref{tab:noise_ratio} shows that our method exhibits strong robustness: even with 50\% noisy data, success rate only drops to 0.50 from 0.73, while efficiency metrics (ASAT and ASST) remain stable. Performance degrades gradually as noise increases. This robustness stems from GRPO's group-relative advantage computation, which effectively filters suboptimal demonstrations during optimization. These results indicate that our single-turn GRPO approach can successfully learn from imperfect expert trajectories, valuable for practical deployment where obtaining perfectly optimal demonstrations is challenging.
}

\section{Conclusion}
\textcolor{black}{In this paper, we present a theoretical framework demonstrating that GRPO improvements on single-turn task reasoning provide a lower bound on multi-turn success probability under minimal steps for challenging multi-turn interaction scenarios.} Empirical results validate our approach, with our small-parameter model consistently outperforming larger baselines while requiring fewer steps for task completion. We also show cross-task generalization capabilities both theoretically and empirically.
While limitations remain regarding the reliance on expert trajectories, our work shows how to leverage single-turn RL successes for complex multi-turn task planning, opening promising directions for more capable and efficient LLM-based agents.

\bibliographystyle{ieeetr}

\bibliography{arxiv}

\begin{thebibliography}{10}

\bibitem{luo2025large}
J.~Luo, W.~Zhang, Y.~Yuan, Y.~Zhao, {\em et~al.}, ``Large language model agent: A survey on methodology, applications and challenges,'' {\em arXiv preprint arXiv:2503.21460}, 2025.

\bibitem{xi2025rise}
Z.~Xi, W.~Chen, X.~Guo, W.~He, Y.~Ding, B.~Hong, M.~Zhang, J.~Wang, S.~Jin, E.~Zhou, {\em et~al.}, ``The rise and potential of large language model based agents: A survey,'' {\em Science China Information Sciences}, vol.~68, no.~2, p.~121101, 2025.

\bibitem{erdoganplan}
L.~E. Erdogan, H.~Furuta, S.~Kim, N.~Lee, S.~Moon, G.~Anumanchipalli, K.~Keutzer, and A.~Gholami, ``Plan-and-act: Improving planning of agents for long-horizon tasks,'' in {\em Forty-second International Conference on Machine Learning}, 2025.

\bibitem{yao2023react}
S.~Yao, J.~Zhao, D.~Yu, N.~Du, I.~Shafran, K.~Narasimhan, and Y.~Cao, ``React: Synergizing reasoning and acting in language models,'' in {\em International Conference on Learning Representations (ICLR)}, 2023.

\bibitem{shinn2023reflexion}
N.~Shinn, F.~Cassano, A.~Gopinath, K.~Narasimhan, and S.~Yao, ``Reflexion: Language agents with verbal reinforcement learning,'' {\em Advances in Neural Information Processing Systems}, vol.~36, pp.~8634--8652, 2023.

\bibitem{zhou2024language}
A.~Zhou, K.~Yan, M.~Shlapentokh-Rothman, H.~Wang, and Y.-X. Wang, ``Language agent tree search unifies reasoning, acting, and planning in language models,'' in {\em Proceedings of the 41st International Conference on Machine Learning}, pp.~62138--62160, 2024.

\bibitem{guan2025evaluating}
S.~Guan, H.~Xiong, J.~Wang, J.~Bian, B.~Zhu, and J.-g. Lou, ``Evaluating llm-based agents for multi-turn conversations: A survey,'' {\em arXiv preprint arXiv:2503.22458}, 2025.

\bibitem{zhou2025sweet}
Y.~Zhou, S.~Jiang, Y.~Tian, J.~Weston, S.~Levine, S.~Sukhbaatar, and X.~Li, ``Sweet-rl: Training multi-turn llm agents on collaborative reasoning tasks,'' {\em arXiv preprint arXiv:2503.15478}, 2025.

\bibitem{wang2025ragen}
Z.~Wang, K.~Wang, Q.~Wang, P.~Zhang, L.~Li, Z.~Yang, X.~Jin, K.~Yu, M.~N. Nguyen, L.~Liu, {\em et~al.}, ``Ragen: Understanding self-evolution in llm agents via multi-turn reinforcement learning,'' {\em arXiv preprint arXiv:2504.20073}, 2025.

\bibitem{feng2025group}
L.~Feng, Z.~Xue, T.~Liu, and B.~An, ``Group-in-group policy optimization for llm agent training,'' {\em arXiv preprint arXiv:2505.10978}, 2025.

\bibitem{dong2025agentic}
G.~Dong, H.~Mao, K.~Ma, L.~Bao, Y.~Chen, Z.~Wang, Z.~Chen, J.~Du, H.~Wang, F.~Zhang, {\em et~al.}, ``Agentic reinforced policy optimization,'' {\em arXiv preprint arXiv:2507.19849}, 2025.

\bibitem{cao2025skyrl}
S.~Cao, S.~Hegde, D.~Li, T.~Griggs, S.~Liu, E.~Tang, J.~Pan, X.~Wang, A.~Malik, G.~Neubig, K.~Hakhamaneshi, R.~Liaw, P.~Moritz, M.~Zaharia, J.~E. Gonzalez, and I.~Stoica, ``Skyrl-v0: Train real-world long-horizon agents via reinforcement learning,'' 2025.

\bibitem{jin2025search}
B.~Jin, H.~Zeng, Z.~Yue, J.~Yoon, S.~Arik, D.~Wang, H.~Zamani, and J.~Han, ``Search-r1: Training llms to reason and leverage search engines with reinforcement learning,'' {\em arXiv preprint arXiv:2503.09516}, 2025.

\bibitem{qi2024webrl}
Z.~Qi, X.~Liu, I.~L. Iong, H.~Lai, X.~Sun, W.~Zhao, Y.~Yang, X.~Yang, J.~Sun, S.~Yao, {\em et~al.}, ``Webrl: Training llm web agents via self-evolving online curriculum reinforcement learning,'' {\em arXiv preprint arXiv:2411.02337}, 2024.

\bibitem{xi2025agentgym}
Z.~Xi, J.~Huang, C.~Liao, B.~Huang, H.~Guo, J.~Liu, R.~Zheng, J.~Ye, J.~Zhang, W.~Chen, {\em et~al.}, ``Agentgym-rl: Training llm agents for long-horizon decision making through multi-turn reinforcement learning,'' {\em arXiv preprint arXiv:2509.08755}, 2025.

\bibitem{peiyuan2024agile}
F.~Peiyuan, Y.~He, G.~Huang, Y.~Lin, H.~Zhang, Y.~Zhang, and H.~Li, ``Agile: A novel reinforcement learning framework of llm agents,'' {\em Advances in Neural Information Processing Systems}, vol.~37, pp.~5244--5284, 2024.

\bibitem{shao2024deepseekmath}
Z.~Shao, P.~Wang, Q.~Zhu, R.~Xu, J.~Song, X.~Bi, H.~Zhang, M.~Zhang, Y.~Li, Y.~Wu, {\em et~al.}, ``Deepseekmath: Pushing the limits of mathematical reasoning in open language models,'' {\em arXiv preprint arXiv:2402.03300}, 2024.

\bibitem{guo2025deepseek}
D.~Guo, D.~Yang, H.~Zhang, J.~Song, R.~Zhang, R.~Xu, Q.~Zhu, S.~Ma, P.~Wang, X.~Bi, {\em et~al.}, ``Deepseek-r1: Incentivizing reasoning capability in llms via reinforcement learning,'' {\em arXiv preprint arXiv:2501.12948}, 2025.

\bibitem{mroueh2025reinforcement}
Y.~Mroueh, ``Reinforcement learning with verifiable rewards: Grpo's effective loss, dynamics, and success amplification,'' {\em arXiv preprint arXiv:2503.06639}, 2025.

\bibitem{gonzalezrobotouille}
G.~Gonzalez-Pumariega, L.~S. Yean, N.~Sunkara, and S.~Choudhury, ``Robotouille: An asynchronous planning benchmark for llm agents,'' in {\em The Thirteenth International Conference on Learning Representations}.

\bibitem{dubey2024llama}
A.~Dubey, A.~Jauhri, A.~Pandey, A.~Kadian, {\em et~al.}, ``The llama 3 herd of models,'' {\em arXiv e-prints}, pp.~arXiv--2407, 2024.

\bibitem{sheng2025hybridflow}
G.~Sheng, C.~Zhang, Z.~Ye, X.~Wu, W.~Zhang, R.~Zhang, Y.~Peng, H.~Lin, and C.~Wu, ``Hybridflow: A flexible and efficient rlhf framework,'' in {\em Proceedings of the Twentieth European Conference on Computer Systems}, pp.~1279--1297, 2025.

\bibitem{qwen2.5}
A.~Yang, B.~Yang, B.~Zhang, B.~Hui, B.~Zheng, B.~Yu, and et~al, ``Qwen2.5 technical report,'' {\em arXiv preprint arXiv:2412.15115}, 2024.

\bibitem{schulman2017proximal}
J.~Schulman, F.~Wolski, P.~Dhariwal, A.~Radford, and O.~Klimov, ``Proximal policy optimization algorithms,'' {\em arXiv preprint arXiv:1707.06347}, 2017.

\bibitem{ahmadian2024back}
A.~Ahmadian, C.~Cremer, M.~Gall{\'e}, M.~Fadaee, J.~Kreutzer, O.~Pietquin, A.~{\"U}st{\"u}n, and S.~Hooker, ``Back to basics: Revisiting reinforce style optimization for learning from human feedback in llms,'' {\em arXiv preprint arXiv:2402.14740}, 2024.

\bibitem{hu2025reinforce++}
J.~Hu, J.~K. Liu, H.~Xu, and W.~Shen, ``Reinforce++: An efficient rlhf algorithm with robustness to both prompt and reward models,'' {\em arXiv preprint arXiv:2501.03262}, 2025.

\end{thebibliography}

\end{document}